%% file: ISCAS.tex
\documentclass[conference]{IEEEtran}
\IEEEoverridecommandlockouts
\usepackage{cite}
\usepackage{amsmath,amssymb,amsfonts}
\usepackage{algorithmic}
\usepackage{algorithm}
\usepackage{graphicx}
\usepackage{textcomp}
\usepackage{xcolor}
\usepackage{caption}
\usepackage{subcaption}
\usepackage{dsfont}
\usepackage{tabularx}
\usepackage{multirow}
\usepackage{hyperref}
\usepackage{bbm}
\usepackage{array, makecell}

\usepackage[compact]{titlesec}
\titlespacing{\section}{0pt}{*1}{*1}

\newcolumntype{C}{>{\centering\arraybackslash}X}
\def\BibTeX{{\rm B\kern-.05em{\sc i\kern-.025em b}\kern-.08em
    T\kern-.1667em\lower.7ex\hbox{E}\kern-.125emX}}

\newcommand{\ie}{\textit{i}.\textit{e}.}
\definecolor{blue(ncs)}{rgb}{0.0, 0.53, 0.74}
\definecolor{applegreen}{rgb}{0.55, 0.71, 0.0}
\begin{document}


\title{Histogram-Equalized Quantization for logic-gated Residual Neural Networks\\

}
\author{\IEEEauthorblockN{Van Thien Nguyen, William Guicquero and Gilles Sicard}
CEA-LETI, F-38000, Grenoble, France [Email: vanthien.nguyen@cea.fr]}

\makeatletter
\def\ps@IEEEtitlepagestyle{
  \def\@oddfoot{\mycopyrightnotice}
  \def\@evenfoot{}
}
\def\mycopyrightnotice{
  {\footnotesize
  \begin{minipage}{\textwidth}
  \centering
  Copyright~\copyright~2022 IEEE. Personal use of this material is permitted. However, permission to use this material \\ 
  for any other purposes must be obtained from the IEEE by sending an email to pubs-permissions@ieee.org. DOI: \href{https://doi.org/10.1109/ISCAS48785.2022.9937290}{10.1109/ISCAS48785.2022.9937290}
  \end{minipage}
  }
}

\maketitle

\begin{abstract}
Adjusting the quantization according to the data or to the model loss seems mandatory to enable a high accuracy in the context of quantized neural networks. This work presents Histogram-Equalized Quantization (HEQ), an adaptive framework for linear symmetric quantization. HEQ automatically adapts the quantization thresholds using a unique step size optimization. We empirically show that HEQ achieves state-of-the-art performances on CIFAR-10. Experiments on the STL-10 dataset even show that HEQ enables a proper training of our proposed logic-gated (OR, MUX) residual networks with a higher accuracy at a lower hardware complexity than previous work.
\end{abstract}


\section{Introduction}
Designing low-precision networks \cite{qnn_2016} is a promising area of research aiming at reducing the bit width to represent weights and activations, thus reducing the overall computational complexity and memory-related costs, namely for performing inference at the edge. The advantages of quantization have been demonstrated on several resource-efficient low-precision CNN accelerators \cite{lee_unpu_2019}, \cite{knag_617_2020}, \cite{kim_resource-efficient_2021}, \cite{Andri2021ChewBaccaNNAF}, \cite{9180844}. 

Quantization-Aware Training (QAT) is the common approach to preserve the performance of quantized models and avoid unacceptable accuracy degradation due to the limited precision. QAT usually consists in using real-valued proxies of the model weights that are on-the-fly quantized during the forward pass while being updated during the backward pass \cite{Courbariaux2016BinarizedNN}. Although several nonlinear quantization mappings \cite{Han2016DeepCC}, \cite{Miyashita2016ConvolutionalNN} and \cite{Zhang2018LQNetsLQ} have demonstrated remarkable algorithmic performances, they are not fully compliant with a simple hardware implementation. On the contrary, a linear symmetric mapping \cite{Zhao2020LinearSQ} naturally matches a streamlined hardware, making it a more relevant and reasonable choice for model quantization.

In the case of linear symmetric quantization, the thresholds are derived from a unique step size. The calibration of this scaling factor plays a key role and we state that it is likely intractable to find the optimal a priori value, given that it deeply depends on the model topology, its initialization, the inference task and the training procedure. Therefore, using an adjustable scaling factor during the training has demonstrated to be more favorable because of taking into consideration the evolution of weight/activation layer-wise distributions. State-of-the-art methods propose to optimize these parameters under the minimization of the quantization error and/or the loss function. For example, \cite{Li2016TernaryWN} aims at minimizing the mean squared error between the floating-point weights and their ternarization while \cite{Zhao2020LinearSQ} updates the step size through a simulated gradient in which the descent direction is based on the quantization error. On the other hand, \cite{ZhuHMD17} and later \cite{li_trq_2021} propose to learn the scaling factor using the task loss backpropagation. Furthermore, \cite{EsserMBAM20} introduces a Straight-Through-Estimated (STE\cite{bengio_estimating_2013}) gradient of the step size with respect to the loss. Based on this work, \cite{UhlichMCYGTKN20} and \cite{nakata_adaptive_2021} take advantage of bitwidth-dependent regularizations to optimize the layer-wise bit allocation given a target model size or a computational budget.

In this paper, we claim that --in most cases-- a proper quantization scheme should cover all the available data representation space, somehow maximizing the entropy of the weights \cite{shannon}. Based on this hypothesis, \cite{He2016EffectiveQM} presents a 2-bit quantization methods for recurrent models where the step size equals to a constant multiple of the mean value of the proxy weights. Similarly, \cite{alom_effective_2018} determines the thresholds of 3-value and 4-value quantizations according to the mean and the standard deviation of the proxy weights. However, these approaches are not generic and applied only to under 3-bit quantization. On the contrary, our proposed method --Histogram-Equalized Quantization (HEQ)-- automatically adjusts the step size of an $n$-value quantization according to its $n$-quantiles such that the resulting quantized values are more balanced, without any further regularization. We empirically show that our method provides a better accuracy than previous methods on different topology variants, from the baseline plain model to our proposed logic-gated residual networks. Indeed, HEQ advantageously enables a proper training of quantized models that embed OR and MUX logic gates to replace floating-point types of skip connections, this way simplifying the Hardware mapping of layer interconnections.

\section{Linear symmetric quantization}
This paper focuses on the linear symmetric quantization to a restricted range of odd $n>2$ discrete values. We consider the mapping $g:\mathbb{R} \rightarrow [[-1,+1]]$ applied to the weight $w$ as:
\begin{equation}
g(w; s) = \frac{2}{n-1} \textrm{Clip}\left( \left \lfloor\frac{w}{s} \right \rceil, \frac{1-n}{2}, \frac{n-1}{2} \right),
\label{linear_symmetric_quant}
\end{equation}
where $[[-1,+1]]$ is discretized with an output step size of $\frac{2}{n-1}$, $s$ is the input step size and $\textrm{Clip}(x; a,b) = \textrm{min}(\textrm{max}(x, a), b)$ with $a < b$. While existing works usually keep the range of floating values by the factor $s$ outside of the clipping function, here we use the $\frac{2}{n-1}$ scale factor so that all the quantized values are explicitely shrunk in the interval $[-1, +1]$. During backpropagation, we use the straight-through-estimated (STE\cite{bengio_estimating_2013}) gradient $\frac{\partial g}{\partial w} = \mathds{1}_{\{|x|\leq 1\}}$ to update the proxy weights. 

\begin{figure}
    \centering
    \hspace{-6mm}
    \begin{subfigure}[b]{0.18\textwidth}
        \includegraphics[width=\textwidth]{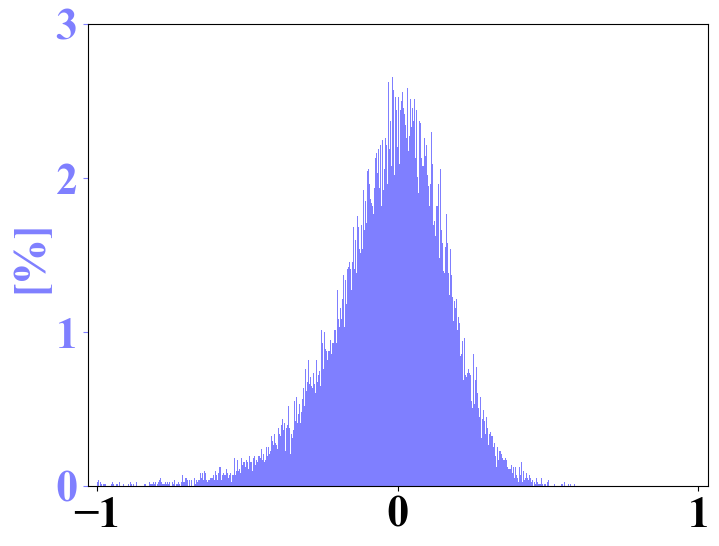}
        \vspace{-6mm}
        \caption{Full-precision model.}
        \label{layer1}
    \end{subfigure}
    \hspace{2mm}
    \begin{subfigure}[b]{0.18\textwidth}
        \includegraphics[width=\textwidth]{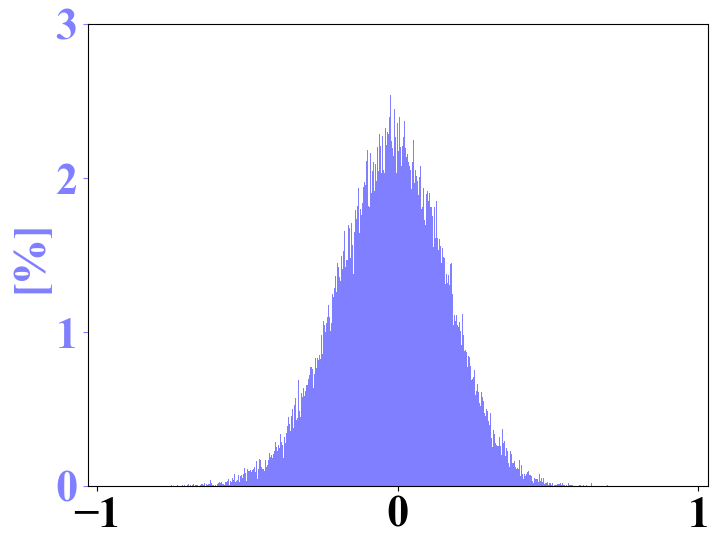}
        \vspace{-6mm}
        \caption{Full-precision model.}
        \label{layer2}
    \end{subfigure}
    
    ~
    \hspace{-4mm}
    \begin{subfigure}[b]{0.2\textwidth}
        \includegraphics[width=\textwidth]{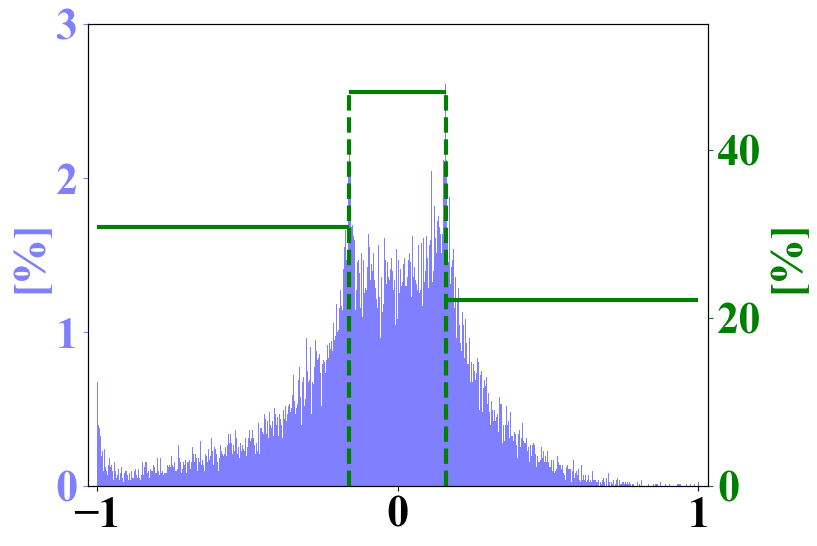}
        \vspace{-6mm}
        \caption{TWN-Ternary weights}
        \label{twn}
    \end{subfigure}
    \begin{subfigure}[b]{0.2\textwidth}
        \includegraphics[width=\textwidth]{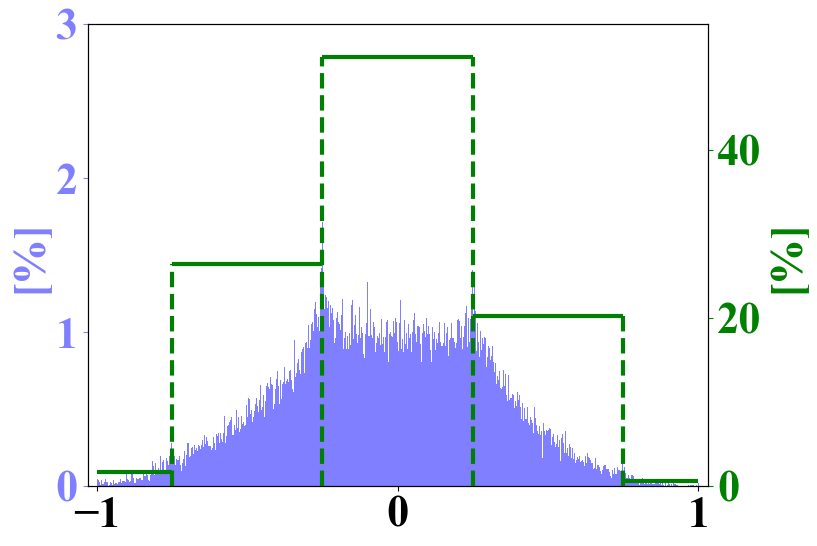}
        \vspace{-6mm}
        \caption{DoReFa-quinary weights}
        \label{dorefa-quinary}
    \end{subfigure}
    ~
     \begin{subfigure}[b]{0.2\textwidth}
        \includegraphics[width=\textwidth]{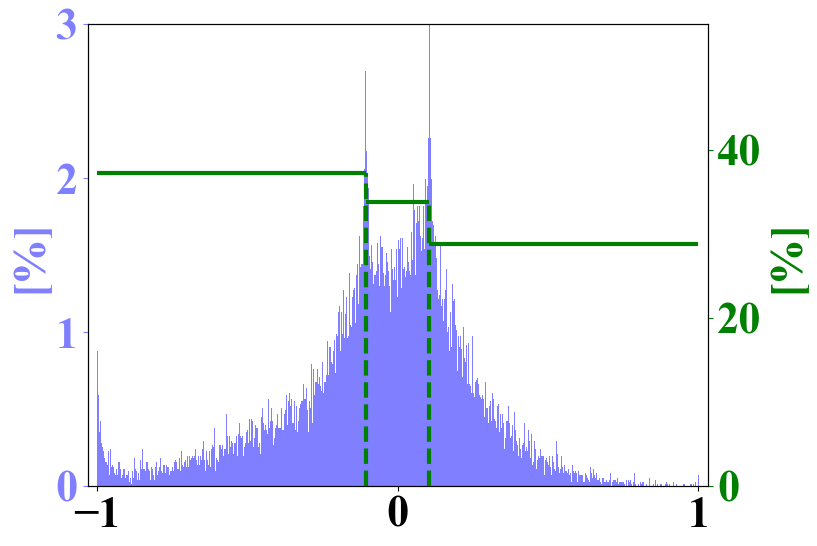}
        \vspace{-6mm}
        \caption{HEQ-ternary weights}
        \label{heq3}
    \end{subfigure}
    \begin{subfigure}[b]{0.2\textwidth}
        \includegraphics[width=\textwidth]{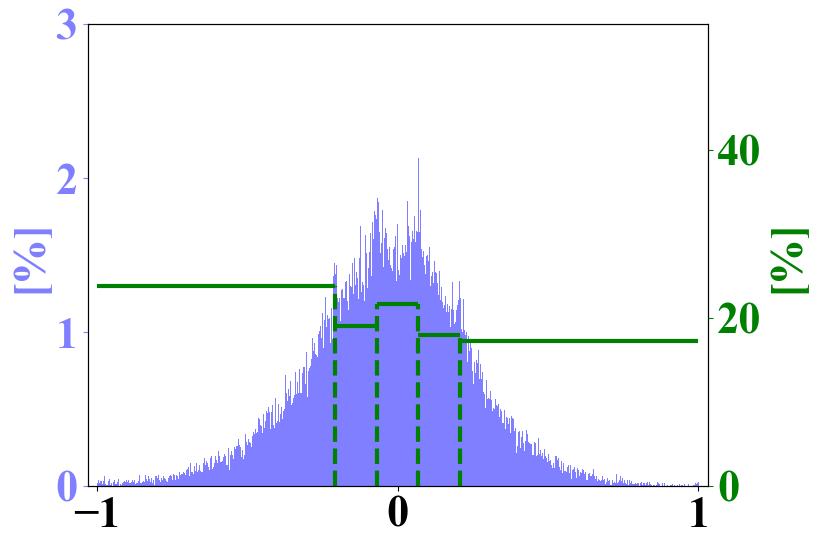}
        \vspace{-6mm}
        \caption{HEQ-quinary weights}
        \label{heq5}
    \end{subfigure}
    \caption{Weight distributions of 2 layers (in 2 columns) after training of: full-precision model ($1^{st}$ line), existing ternary-weight and quinary-weight model ($2^{nd}$ line), and our proposed HEQ method ($3^{rd}$ line) along with quantization thresholds.}
    \label{proxy_hist}
    \vspace{-4mm}
\end{figure}

This formulation thus depends on the definition of $s$ that deeply impacts on the model accuracy. Let us consider Ternary Weight Networks (TWN)\cite{Li2016TernaryWN} as the baseline where $s = 2\tau \frac{\sum_{i=1}^{n} |W_i|}{n}$, with a fixed norm factor $\tau = 0.7$. Although the optimal $s$ may change depending on the data distribution, this method cannot be applied to higher precisions and the predefined $\tau$ limits the adaptability of the quantizer. Similarly, DoReFa\cite{Zhou2016DoReFaNetTL} forces the real values into the range of $[-1, +1]$ by a mapping adapted to the data, but the thresholds remain fixed. Fig.~\ref{proxy_hist} depicts the histogram of proxy weights (blue bars) and quantized weights (horizontal green lines) in the case of 3-value (Figs.~\ref{twn},~\ref{heq3}) and 5-value (Figs.~\ref{dorefa-quinary},~\ref{heq5}) quantization whose initialization (from full-precision model) is shown in Figs.~\ref{layer1} and~\ref{layer2}, respectively. We can observe that in both cases of TWN (Fig.~\ref{twn}) and DoReFa (Fig.~\ref{dorefa-quinary}), the proxy weights are mainly concentrated around zero and the distributions between thresholds (vertical green lines) are unbalanced. In particular, the quinary weights (3-bit) in Fig.~\ref{dorefa-quinary} can be approximated by only 3 values (2-bit). Consequently, the quantized weights fail to exploit all available values which may cause the model to be sub-optimal. This motivates the use of more proper quantizers, favoring the balance of quantized weights.

\section{Histogram-Equalized Quantization (HEQ)}
To resolve the aforementioned imbalance between quantized values, we propose HEQ to automatically adjust $s$ during training. Assuming that a proper quantizer should optimize the balanced use of available discrete values in the data representation space, we iteratively tune $s$ based on the histogram of the proxy weights to equi-distribute quantized weights. 

\subsection{Method}
\begin{figure}[h]
	\centerline{\includegraphics[scale=0.25]{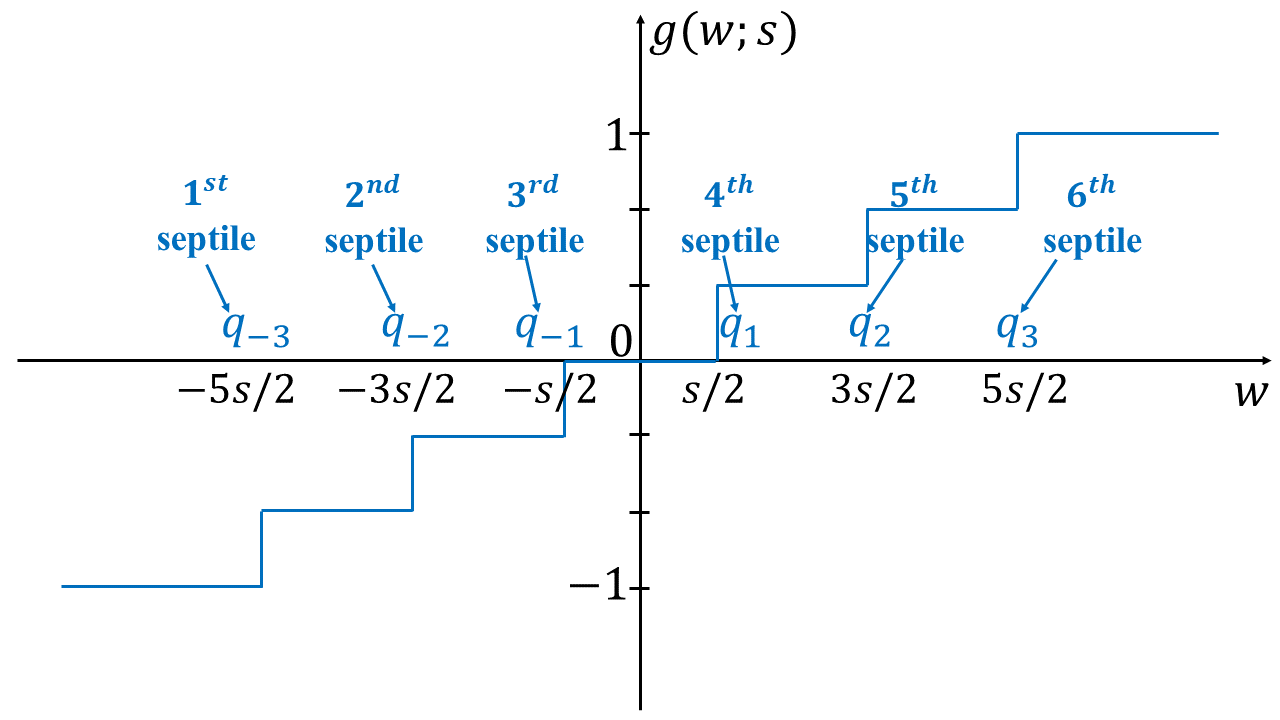}}
    \caption{Symmetric linear quantization with histogram bin equalization when $n$-quantiles ($q_{-i}, q_i$) are symmetrical and coincide with the quantized thresholds.} 
	\label{linear_quant}
\end{figure}

In Fig.~\ref{linear_quant} we denote $\{(q_i, q_{-i})\}_{i \in [\![1,\frac{n-1}{2}]\!]}$ as $n-1$ points which divide the histogram of weights into $n$ intervals with equiprobabilities (namely $n$-quantiles). Observing that the weights distribution may change during the training procedure but with a median value that usually stays around zero, we assume that these quantiles are symmetrically distributed around zero, \ie \, $q_{-i} \approx - q_i$ with $q_i > 0$.

In order to equalize the histogram bins of quantized values, we thus re-estimate and update $s$ such that the resulting thresholds used by the quantization function (see Fig.~\ref{linear_quant}) are getting closer to these quantiles. Therefore, $s$ can be approximated by a weighted sum of the quantiles such that $q_i$ approximately coincides with $\frac{(2i-1)s}{2}$. We thus propose to update $s$ using the following approximation given the aforementioned symmetry:

\begin{equation}
\sum_{i=1}^{\frac{n-1}{2}} (|q_{-i}| + q_i) = 2 \sum_{i=1}^{\frac{n-1}{2}} \frac{(2i-1)s}{2} ,
\label{stepsize_approx}
\end{equation}

from which we can derive the following updating formula:

\begin{equation}
s = \frac{4\sum_{i=1}^{\frac{n-1}{2}} (|q_{-i}| + q_i)}{(n-1)^2}.
\label{stepsize}
\end{equation}
This approach has the great advantage of being generic, compatible with almost all use cases regardless the quantization level, the position of the layer and its type (with a possible extension to an even $n$). To maintain the stability of the model during optimization, we compute and update $s$ only at the beginning of each epoch. The formal training procedure is detailed in Algorithm~\ref{algo}. The proxy weight distributions obtained after a training stage that are reported in Figs.~\ref{heq3} and ~\ref{heq5} clearly demonstrate that HEQ provides a more balanced distribution of quantized weights.

Note that with symmetric quantiles, the resulting quantized weights become more equalized. This can be enhanced by forcing the weight median to zero, which has not been applied in the scope of this work for the sake of clarity since it seems to have only a slight impact on the performance of the model.

\begin{algorithm}
 \caption{Training QNN with Histogram-Equalized Quantization (HEQ)}
 \begin{algorithmic}[1]
 \label{algo}
 \renewcommand{\algorithmicrequire}{\textbf{Input:}}
 \renewcommand{\algorithmicensure}{\textbf{Output:}}
\REQUIRE Initial proxy weights $\{\textbf{W}_l\}_{l=1}^L$ and training dataset
 \ENSURE  Optimized $\{\textbf{W}_l\}_{l=1}^L$,  $\{s_l\}_{l=1}^L$ \\
 // $B$, $I$, $L$: \#batches, \#epochs, \#layers\\
 // $\textbf{W}_l$: full-precision proxy weights of the $l^{th}$ layer,\\
 // $s_l$: quantization step size used at the $l^{th}$ layer. \\
  \FOR {$i = 1$ to $I$} 
  \FOR {$l = 1$ to $L$}
  \STATE Find $n$-quantiles of layer $l$
  \STATE Compute and update $s_l$ (Eq.~\ref{stepsize})
  \ENDFOR
  \FOR{$b = 1$ to $B$}
  \STATE Forward pass using $\{g(\textbf{W}_l; s_l)\}_{l=1}^L$ (Eq.~\ref{linear_symmetric_quant})
  \STATE Backward pass and update $\{\textbf{W}_l\}_{l=1}^L$
  \ENDFOR
  \ENDFOR
 \end{algorithmic} 
 \end{algorithm}

\subsection{State-of-the-art benchmark} \label{sota_benchmark}
\begin{figure}[h]
     \centering
         \includegraphics[width=0.38\textwidth]{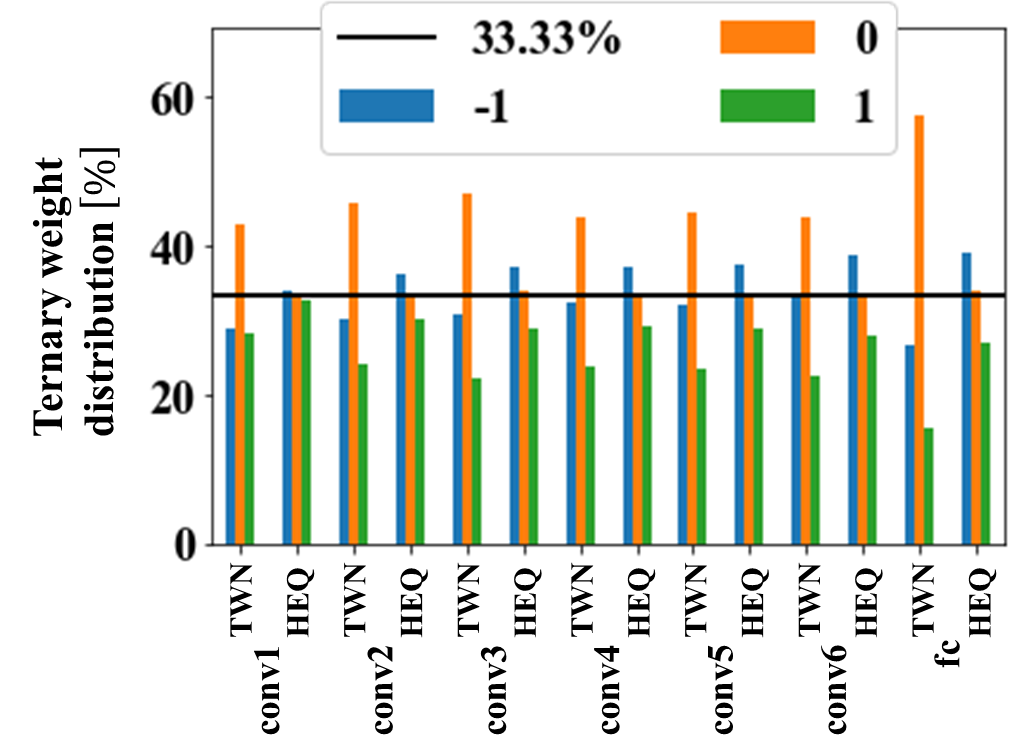}
         \caption{Comparison of the ternary-weight distribution using TWN and our HEQ method.}
         \label{twn_hist}
         \vspace{-3mm}
\end{figure}

HEQ has been evaluated on the CIFAR-10 dataset\cite{krizhevsky_learning_nodate} with $32\times32$ RGB images and using the VGG-Small model like in \cite{Zhang2018LQNetsLQ}. A combination of a scale-invariant random crop (performed on all-sided 4-pixel padded images) combined with a random horizontal flip is used for data augmentation. Initial proxy weights are from a pre-trained full-precision network. Motivated by Hardware considerations, a 2-bit activation scheme as detailed in DoReFa \cite{Zhou2016DoReFaNetTL} has been used. Our model is trained during 100 epochs with a small batch size of 50 to favor exploration. The learning rate is set to $10^{-3}$ during the first 50 epochs, then exponentially rescaled by a factor of $0.9$ at each epoch. Finally, a very last epoch with a larger batch size of 100 and a smaller learning rate of $10^{-5}$ is performed for fine-tuning. Fig.~\ref{twn_hist} allows a comparison between TWN\cite{Li2016TernaryWN} and our HEQ method with respect to the resulting weight distributions. While the zero values dominate all layers in the case of TWN, our method reduces the variance and limits the number of weights at 0 to nearly $1/3$ as shown in Fig.~\ref{twn_hist}. The number of $-1$ values slightly dominates as more proxy weights are concentrated on the negative side.

\begin{figure}[h]
     \centering
    \begin{subfigure}[b]{0.4\textwidth}
        \includegraphics[width=\textwidth]{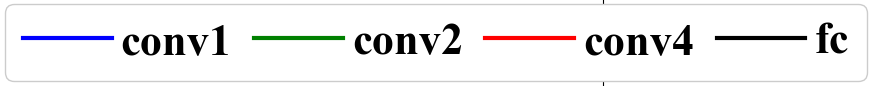}
    \end{subfigure}
         \begin{subfigure}[b]{0.24\textwidth}
        \includegraphics[width=\textwidth]{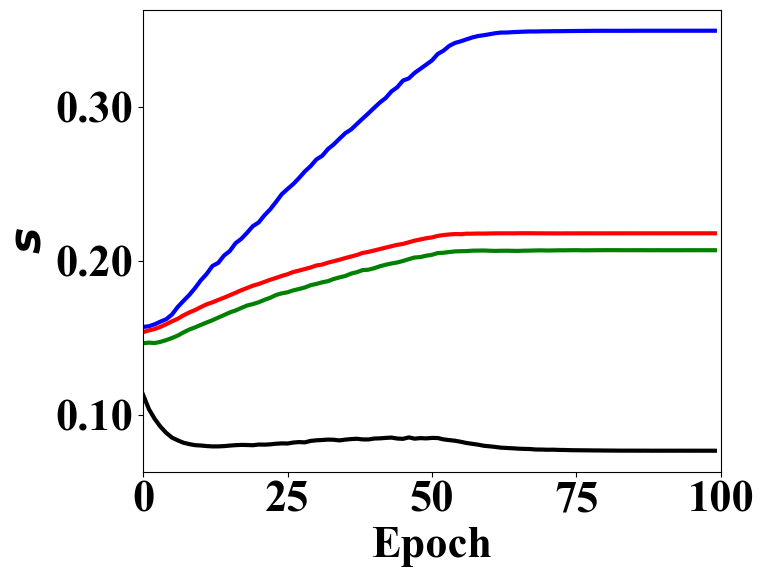}
        \caption{Ternary quantization.}
        \label{ternary}
    \end{subfigure}
    \hspace{-2mm}
    \begin{subfigure}[b]{0.24\textwidth}
        \includegraphics[width=\textwidth]{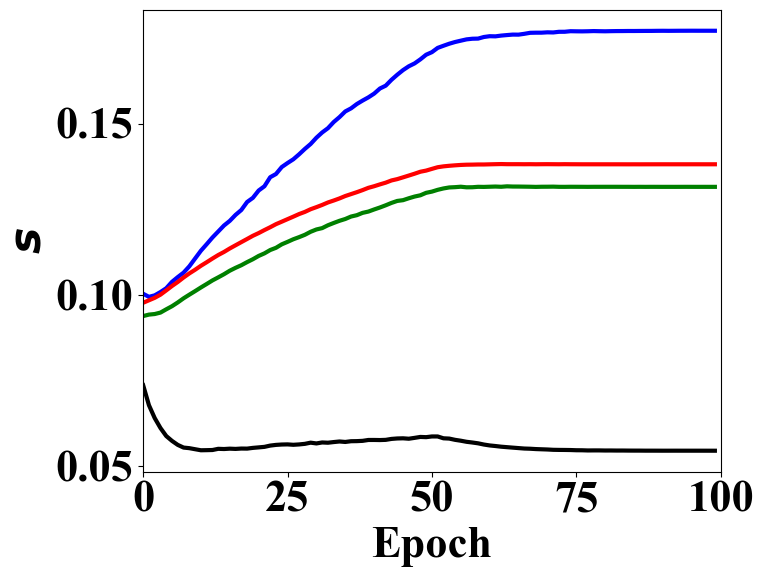}
        \caption{Quinary quantization.}
        \label{quinary}
    \end{subfigure}
     \caption{Evolution of the step size $s$ during training.}
     \label{stepsize_curves}
\end{figure}

Fig.~\ref{stepsize_curves} depicts the variation of $s$ during training in both ternary and quinary cases. It shows that the evolution of $s$ depends on the layer and has different convergence values.

Table~\ref{sota_table} reporting the average accuracy of each configuration over 5 realizations, demonstrates the competitiveness of HEQ compared to the state-of-the-art quantization methods. For instance, when quantizing both weights (W) and activations (A) into 2-bit, we obtain 93.51$\%$ accuracy while having only 3 values $\{-1, 0, +1\}$ over 4 values possible like LQ\cite{Zhang2018LQNetsLQ} and LLSQ\cite{Zhao2020LinearSQ}. Compared to the full-precision model, we observe mostly no degradation in the case of quinary weights ($n=5$) and even a gain with septenary weights ($n=7$). Moreover, while other works give rise to a full-precision scaling factor besides the integer weights which demands the fusion into Batch Normalization \cite{Ioffe2015BatchNA} (BN) for later hardware implementation, our models trained with HEQ-ternary and HEQ-quinary obtain directly integer values $0, \pm 1$ (logical operations) and $\pm 0.5$ (bitshifts) which is already compatible for an easy hardware deployment. 

\begin{table}[h]
\caption{Comparison with the state-of-the-art low-precision quantization methods on CIFAR-10.}
\centering
\begin{tabular}{| c | c | c | c |}
\hline
Method & HW-Compatibility & Bitwidth W/A & Accuracy($\%$) \\
\hline
TWN\cite{Li2016TernaryWN}   & +   & 2/32  & 92.56 \\ \hline
STTN \cite{Xu2020SoftTT}   & +   & 2/2  & 92.93 \\ \hline
TRQ \cite{li_trq_2021}    & +   & 2/2  & 91.2  \\ \hline
\multirow{2}{*}{LQ\cite{Zhang2018LQNetsLQ}}   & \multirow{2}{*}{-}   & 2/32 & 93.8  \\ 
       &   & 2/2  & 93.50 \\ \hline
LLSQ \cite{Zhao2020LinearSQ}  & +   & 2/2  & 93.31 \\ \hline
FP32 baseline   &    & 32/32 & 93.68 \\
\textbf{HEQ-ternary} & ++ & 2/2 & 93.51 \\ 
\textbf{HEQ-quinary} & ++ & 3/2 & 93.66 \\ 
\textbf{HEQ-septenary} & + & 3/2 & 93.75 \\ \hline
\end{tabular}
\label{sota_table}
\end{table}
\section{Extensions to binarized skip connections}
\subsection{Logic-gated Residual Neural Networks}
Although quantization methods have been mainly applied to a wide range of DNN topologies, their usage is mainly focused on reducing the weight and activation bit widths. On the other hand, the element-wise addition in the case of skip connections (ResNet\cite{He2016DeepRL}) is still performed using a full-precision like in \cite{Liu2018BiRealNE}, \cite{Phan2020MoBiNetAM} and \cite{Yao2021HAWQV3DN}. The reason is that apart from improving feature map reusability, these full-precision additions are mostly used to handle the gradient vanishing and mismatching issues, which seem to be even more crucial in the context of quantized models. However, it results in additional costs with respect to the corresponding hardware implementation. In this section, we focus on the compression of those skip connections, including the residual addition and the attention-like multiplication \cite{Wang2017ResidualAN}, such that these element-wise operations can be implemented by only OR and MUX logic gates rather than 32-bit arithmetic hardware.

\begin{figure}[h]
	\centerline{\includegraphics[scale=0.25]{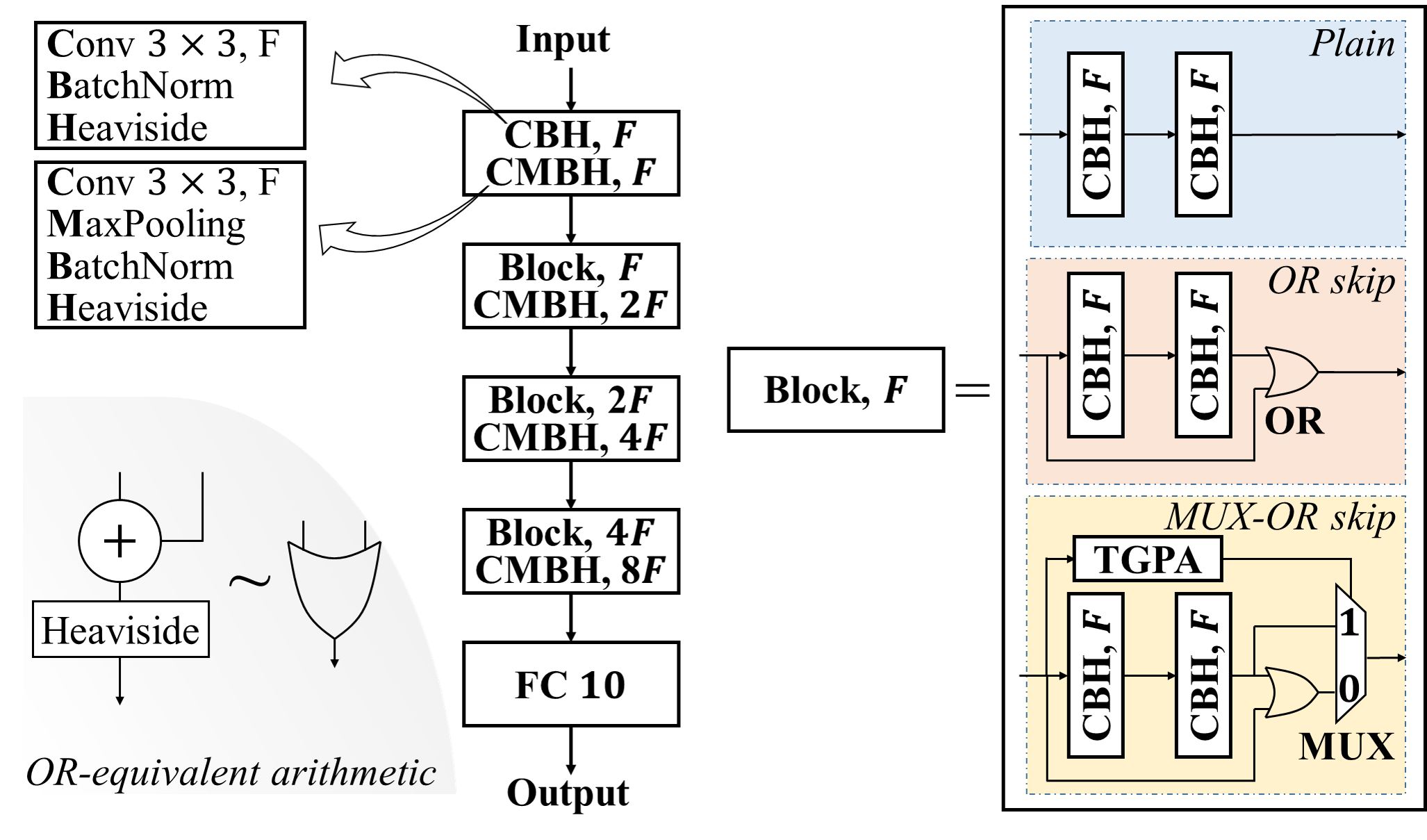}}
	\caption{Models with the plain block (11-hidden layer VGG\cite{VGGNet}-variant), OR-gated block and MUX-OR gated block.} 
	\label{topology}
\end{figure}    

Fig.~\ref{topology} depicts the proposed model design with 3 consecutive convolution blocks with different variants: plain block (denoted as VGG-11), OR block (ORNet-11) and MUX-OR block (MUXORNet-11), where $F$ denotes the basis number of convolutional output feature maps. All the activations are binarized using the Heaviside function $H(x) = \mathds{1}_{\{x > 0\}}$ where $\mathds{1}$ is the indicator function. The logical OR operation between two binary inputs is arithmetically performed as $x_1 \lor x_2 = H(x_1 + x_2)$. The MUX-OR block additionally embeds an attention-alike branch (called MUX branch) along with the OR skip connection. This MUX branch is composed of a channel-wise Thresholded Global Average Pooling (TGPA) that corresponds to a Global Average Pooling (GPA) followed by the (re)binarization $T(x) = \mathds{1}_{\{x > 0.5m\}}$, where $m$ is set to the maximum of the GPA's outputs in full-precision model and to $1$ in quantized model. When deploying the quantized model, this operation can be basically implemented via a bitcount followed by a comparison with a threshold level equal to half the number of pixels. Concretely, the OR-skip connection will be performed for each input feature map channel that has more zeros than ones. Otherwise, the MUX will simply keep the straightforward output of the second Convolution-BatchNorm-Heaviside (CBH) module. One interesting aspect of such a MUX-skip connection is that it favors to balance the number of $1$ with respect to the number of $0$ throughout the networks, this without any other specific regularization. In terms of hardware deployment, while existing approaches with 32-bit additions and multiplications require hundreds of Xilinx FPGA slices \cite{32b_fpga}, a 1-bit OR only costs a single slice and consumes much less energy \cite{Horowitz201411CE}.  

\subsection{Experimental results}

In this section, we evaluate the aforementioned Neural Network topology variants using the proposed HEQ with ternary weights on STL-10 dataset \cite{Coates2011AnAO} of $96\times96$ RGB images. To limit the overfitting, we used the following data augmentation scheme: random crop from all-sided 12-pixel padded images combined with random horizontal flips and cutouts of $32 \times 32$ pixel patches \cite{Devries2017ImprovedRO}. All the parameters of the quantized model are initialized from its pre-trained full-precision network counterpart, in which all Heaviside functions are replaced by ReLU. We set $F=64$ instead of $128$ in VGG-7, resulting in smaller-sized model. All propositions are implemented using Tensorflow \cite{tensorflow2015-whitepaper} and Larq \cite{Geiger2020LarqAO}. 

\begin{table}[h]
\caption{Comparison with the state-of-the-art low-precision quantization methods on STL-10 dataset.}
\centering
\begin{tabular}{|c | c | c | c | c | c | c |}
 \hline
\makecell{Model} & \makecell{Training} & \makecell{Regularization} & \makecell{$\#$ params.\\ (M)} & \makecell{Bitwidth \\ W/A} & \makecell{Acc. \\ ($\%$)}  \\
\hline
\multirow{2}{*}{VGG-7}&  \multirow{2}{*}{LSQ\cite{EsserMBAM20}} & $\#$params$\times$bit \cite{UhlichMCYGTKN20}  & \multirow{2}{*}{4.57} & 2.5/8 & 83.6 \\
 &   & $\#$MACs$\times$bit \cite{nakata_adaptive_2021} &  & 2.2/8 & 83.8  \\
\hline
VGG-11   & \multirow{3}{*}{HEQ} & \multirow{3}{*}{None} & \multirow{3}{*}{3.14} & \multirow{3}{*}{2/1} & 83.34  \\
ORNet-11   &  &  &  &  & 83.82  \\
MUXORNet-11 &  &  & &  & \textbf{84.17} \\
\hline
\end{tabular}
\label{block_comparisons}
\end{table}

Table~\ref{block_comparisons} summarizes the performance of our proposed models compared to previous work\cite{nakata_adaptive_2021} which uses the LSQ\cite{EsserMBAM20} method to jointly adapt the step size and the layer-wise bitwidths under a model size-based \cite{UhlichMCYGTKN20} or a MAC$\times$bit-based regularization. Note that we only took into account the number of convolution parameters for the sake of a fair comparison. While the plain baseline obtains only $83.3\%$ accuracy, ORNet-11 achieves $83.8\%$ and the MUXORNet-11 even achieves up to $84.2\%$, \ie \, a noticeable improvement without increasing the overall model size and with a negligible extra cost of 2-input MUX, OR gates and thresholded bitcounts. In terms of model size, all our $3$ model variants contain less parameters at lower precision compared to \cite{nakata_adaptive_2021}. However, the proposed ORNet-$11$ optimized by HEQ already achieves the same level of accuracy while MUXORNet-$11$ even obtains a better accuracy ($0.37\%$). These results demonstrate the effectiveness of HEQ on different DNN designs, from VGG-like to the proposed ORNet and MUXORNet. It also shows the possibility of compressing the skip connections via logic gates in order to significantly simplify the hardware mapping of more sophisticated ternarized neural network topologies than VGG-like.


\section{Conclusion}
We introduce a novel QAT method based on the equalization of layer-wise weight histograms. During the training process, the step size is adaptively changed according to the proxy weight distribution through its $n$-quantiles, such that the quantized levels are approximately equalized. We empirically show that the models trained with our HEQ can achieve not only state-of-the-art accuracy on CIFAR-10, but even a better accuracy on STL-10 dataset thanks to the proposed logic-gated residual networks, while using a lower precision than
previous works on budget-aware learned quantization.      
\newpage
\newpage
\bibliographystyle{IEEEtran}
\input{ISCAS.bbl}

\end{document}